\documentclass[conference]{IEEEtran}
\IEEEoverridecommandlockouts
\usepackage{cite}
\usepackage{amsmath,amssymb,amsfonts}
\usepackage{algorithmic}
\usepackage{graphicx}
\usepackage{textcomp}
\usepackage{xcolor}
\usepackage{subcaption}
\newcounter{loopc}
\usepackage{fancyhdr}
\NewDocumentCommand\towrite{O{1}m}%
  {{\color{red}#2\forloop{loopc}{1}{\value{loopc} < #1}{; #2}}}

\renewcommand{\vec}[1]{\boldsymbol{#1}}

\NewDocumentCommand{\R}{}{ \mathbb{R} }

\def\BibTeX{{\rm B\kern-.05em{\sc i\kern-.025em b}\kern-.08em
    T\kern-.1667em\lower.7ex\hbox{E}\kern-.125emX}}
\begin{document}

\title{Spatio-Temporal Attention Graph Neural Network for Remaining Useful Life Prediction\\
\thanks{This work is supported within the Digital-Twin-Solar (03EI6024E) project, funded by BMWi: Deutsches Bundesministerium für Wirtschaft und Energie/German Federal Ministry for Economic Affairs and Energy.}
}

\author{\IEEEauthorblockN{1\textsuperscript{st} Zhixin Huang}
\IEEEauthorblockA{\textit{Intelligent Embedded Systems}\\\textit{University of Kassel}\\
Kassel, Germany \\
zhixin.huang@uni-kassel.de}
\and
\IEEEauthorblockN{2\textsuperscript{nd} Yujiang He}
\IEEEauthorblockA{\textit{Intelligent Embedded Systems}\\\textit{University of Kassel}\\
Kassel, Germany \\
yujiang.he@uni-kassel.de}
\and
\IEEEauthorblockN{3\textsuperscript{th} Bernhard Sick}
\IEEEauthorblockA{\textit{Intelligent Embedded Systems}\\\textit{University of Kassel}\\
Kassel, Germany \\
bsick@uni-kassel.de}
}

\maketitle
\thispagestyle{fancy} 
\lhead{This article has been accepted in the International Conference Computational Science \& Computational Intelligence (CSCI'23).} 
\chead{} 
\rhead{} 
\lfoot{} 
\cfoot{} 
\rfoot{\thepage} 
\renewcommand{\headrulewidth}{0pt} 
\renewcommand{\footrulewidth}{0pt} 

\begin{abstract}

Remaining useful life prediction plays a crucial role in the health management of industrial systems. Given the increasing complexity of systems, data-driven predictive models have attracted significant research interest. Upon reviewing the existing literature, it appears that many studies either do not fully integrate both spatial and temporal features or employ only a single attention mechanism. Furthermore, there seems to be inconsistency in the choice of data normalization methods, particularly concerning operating conditions, which might influence predictive performance. To bridge these observations, this study presents the Spatio-Temporal Attention Graph Neural Network. Our model combines graph neural networks and temporal convolutional neural networks for spatial and temporal feature extraction, respectively. The cascade of these extractors, combined with multi-head attention mechanisms for both spatio-temporal dimensions, aims to improve predictive precision and refine model explainability. Comprehensive experiments were conducted on the C-MAPSS dataset to evaluate the impact of unified versus clustering normalization. The findings suggest that our model performs state-of-the-art results using only the unified normalization. Additionally, when dealing with datasets with multiple operating conditions, cluster normalization enhances the performance of our proposed model by up to 27\%.
\end{abstract}

\section{Introduction}
\label{sec:intro}
Predictive Health Management (PHM) stands out as a key tool for reducing maintenance costs and enhancing system reliability and performance in complex industrial systems~\cite{ding2020state}.
PHM systems effectively identify component failures, monitor conditions, and anticipate system breakdowns, assisting or even independently initiating maintenance strategies~\cite{liao2013discovering}.
Remaining Useful Life (RUL) prediction is indispensable for efficacious PHM. The RUL of a system is denoted as the time from the current time to the end of its useful life~\cite{pang2021bayesian}. Precise RUL prognostications facilitate smart system maintenance, preemptively address potential component fault, extend system life, and reduce maintenance costs~\cite{wang2021spatio}.



RUL prediction is mainly divided into physical model-based and data-driven methods~\cite{wang2010trajectory}. The former emphasizes modeling system degradation trajectories using mathematical constructs. However, as the system's complexity increases, it becomes more and more difficult to establish physical models~\cite{jouin2016degradations,pang2021bayesian}. As a result, academic and industrial fields are increasingly leaning towards data-driven approaches, with a particular emphasis on deep learning models~\cite{chen2020machine, yu2020improved,mo2021remaining,li2022emerging,zhu2023rgcnu}. The foundational premise here entails leveraging sensor recording data with their corresponding RULs, utilizing these as input-output pairs, and training a deep neural network. The network reveals complex relationships in the data with its powerful nonlinear modeling capabilities.
\label{subsec:our_method}
\begin{figure*}[htbp]
\centerline{\includegraphics[width=0.9\textwidth]{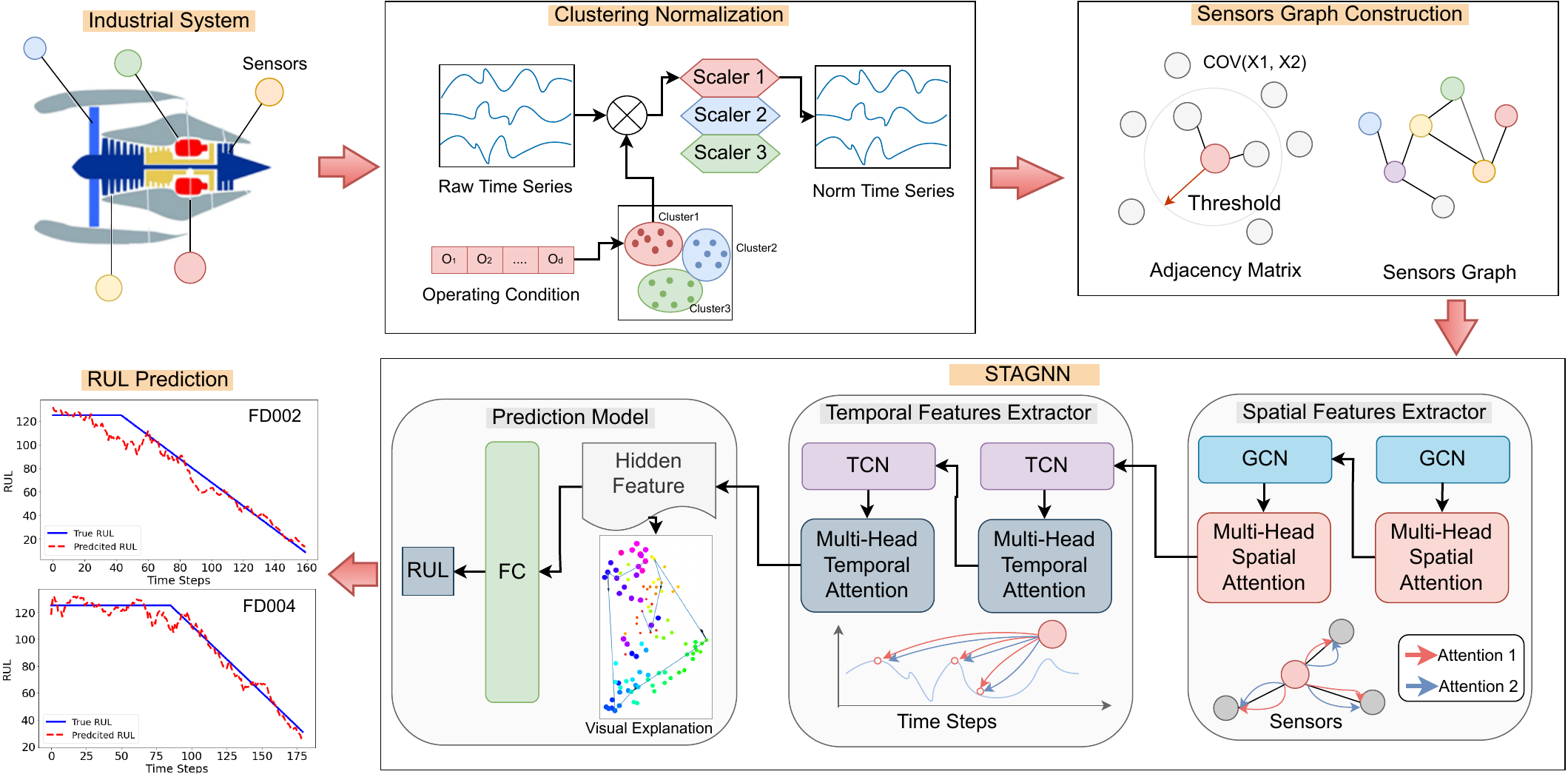}}
\caption{The RUL prediction workflow and the spatio-temporal attention graph neural network}
\label{fig:STAGNN}
\end{figure*}

The data-driven methods encompass Recurrent Neural Networks~(RNN), Convolutional Neural Networks~(CNN), Graph Neural Networks~(GNN), and hybrid models. RNNs, and their advanced version Long Short-Term Memory~(LSTM), are popular for their adeptness in modeling temporal correlations, especially in time series data, rendering them efficacious for RUL predictions~\cite{yu2020improved,liu2020remaining}. Concurrently, CNNs have garnered recognition in RUL prediction~\cite{li2018remaining,zhu2018estimation}, wherein temporal features are extracted via 1D CNN, often resulting in a more effective training process compared to LSTM~\cite{lea2017temporal,liao2018uncertainty, lin2022bayesian}. Beyond the temporal feature, it's imperative to acknowledge the graph-based spatial relations to the multivariate time series engendered by system sensors~\cite{wang2021spatio,li2022emerging}. In addition, hybrid models were proposed to extract spatial and temporal features, respectively by combining GNN and RNN or CNN. spatio-temporal graph neural networks have many applications, such as traffic forecasting~\cite{yu2017spatio}
and activity prediction~\cite{mohamed2020social}. In recent research, \cite{wang2021spatio} and~\cite{zhu2023rgcnu} proposed to apply the spatio-temporal graph neural network to RUL prediction and achieved state-of-the-art~(SOTA) performance. In \cite{wang2021spatio}, graph convolutional network~(GCN)~\cite{kipf2016semi} and temporal convolutional network~(TCN)~\cite{bai2018empirical} were used in parallel to obtain temporal features and spatial features, respectively, and the two features were aggregated through product. Similar to~\cite{wang2021spatio}, Zhu et al~\cite{zhu2023rgcnu} connected GNN and 1D CNN in parallel as the feature extractors.

Despite the advancements in RUL prediction methodologies, current strategies exhibit avenues for further refinement. This research identifies three key questions that will serve as a foundation for subsequent sections, aiming to explore potential enhancements in RUL prediction accuracy.

\textbf{Question 1}: How does cascade integration of spatial and temporal feature extractors affect predictive accuracy? SOTA methods, as discussed in references~\cite{wang2021spatio,zhu2023rgcnu}, predominantly parallelize the extraction of spatio-temporal features, followed by a simple aggregation. An alternative could involve a cascade arrangement where the spatial feature extractor first filters the sensor data based on its graph structure. Subsequently, its output would serve as the input for the temporal feature extractor, thereby allowing a more nuanced feature extraction process.

\textbf{Question 2}: Can introducing a spatio-temporal attention mechanism amplify the prediction's performance? Attention mechanisms have gained traction for their efficacy in highlighting critical information within neural network computations, initially finding success in computer vision and later in natural language processing~\cite{niu2021review}. Its extension to RUL prediction, particularly by integrating spatio-temporal dimensions, remains partially uncharted. Existing studies~\cite{chen2020machine, liu2020remaining, mo2021remaining, li2021hierarchical} typically incorporate only one form of attention, either spatial or temporal. Notably, the SOTA hybrid methodologies ~\cite{wang2021spatio, zhu2023rgcnu} have not yet adopted any form of attention mechanism. Additionally, analyzing the attention probability matrix can provide insight into the information that the predictive model deems most critical.



\textbf{Question 3}: How do varying operating conditions impact predictive accuracy? As an illustration, two subsets of the C-MAPSS dataset~\cite{saxena2008damage} comprise 6 distinct operating conditions. Sensors manifest disparate initial values and variation ranges under these differing conditions. Examining prevailing literature reveals a divergence in data normalization approaches during preprocessing. While certain studies~\cite{miao2019joint,li2021hierarchical,lin2022bayesian} have taken into account the implications of operating conditions on data, but most data-driven methodologies~\cite{li2018remaining,chen2020machine,yu2020improved,liu2020remaining,mo2021remaining,wang2021spatio,kim2020bayesian,zhu2023rgcnu} resort to unified normalization. Unified normalization, as depicted in Fig~\ref{subfig:Old_Normalization}, can inadvertently introduce noise into the normalized dataset. A comparison between Fig.~\ref{subfig:Old_Normalization} and Fig.~\ref{subfig:New_Normalization} suggests that clustering normalization, which factors in operating conditions, might increase prediction accuracy. However, there are no existing works comparing the impacts of divergent data preprocessing methodologies on prediction outcomes. 

The \textbf{contributions} of this paper are delineated into three aspects: 
(1) In addressing questions 1 and 2, we introduce the \textbf{S}patial-\textbf{T}emporal \textbf{A}ttention \textbf{G}raph \textbf{N}eural \textbf{N}etwork~(STAGNN). As depicted in Fig.~\ref{fig:STAGNN}, our model employs GNN and TCN as spatial and temporal extractors, respectively. We seamlessly integrate these two modules through a cascading approach and incorporate a multi-head spatio-temporal attention mechanism, which enhances predictive accuracy and increases model explainability.
(2) In response to question 3, we implement two distinct data preprocessing strategies to assess the influence of operating conditions on predictive outcomes. Beyond the unified normalization, we also introduce the cluster normalization technique. This entails clustering the operating conditions and subsequently normalizing each cluster independently.
(3) Leveraging the C-MAPSS dataset, we conducted a comprehensive experimental evaluation. Our findings demonstrate that, even in the absence of cluster normalization, the proposed STAGNN model consistently achieves SOTA results across all four sub-datasets. Notably, when cluster normalization is incorporated during the data preprocessing phase, there's an augmentation in prediction performance by 26\% in RMSE and 27\% in Score. Furthermore, to shed light on the model's explainability, we provide visualizations of STAGNN's hidden feature representations and the attention matrix.
\section{Methodology}
\label{sec:method}
\subsection{Problem Statement}
\label{subsec:problem_statement}




In industrial systems, sensor networks typically yield data in the form of time series signals. 
For the task of RUL prediction employing data-driven models, deep learning architectures are utilized to extract degradation information from these signals. 
As highlighted in Sec~\ref{sec:intro}, it is imperative to account for the system's operating conditions, which might encompass extreme settings such as high or low temperatures and varying humidity levels. 
Time series data produced under distinct operating conditions can exhibit substantial variation. Therefore, an exclusive focus on sensor data might compromise the model's prediction accuracy.

In the context of this paper, the task of multi-sensor RUL prediction can be defined as: given a multi-sensor time series signal, denoted as $\vec{X}_T$, in conjunction with the system's operating condition $\vec{O}$, we aim to assemble an end-to-end RUL prediction workflow $f( )$, encompassing both data preprocessing and a deep regression model. The multi-sensor signals, given $T$ observational instances, $\vec{X}_T$, is formulated as:
\begin{equation}
\label{eq:X_define}
\vec{X}_T = \vec[\vec{x}_T^1, \vec{x}_T^2, \dots, \vec{x}_T^S] \in \R^{T \times S},
\end{equation}
where $S$ represents the count of sensors. Assuming that the system's operating conditions remain static after startup, each sensor's time series signal $\vec{X}_T$ is generated with a specific operating condition. The vector $\vec{O}$ encompasses $d$ conditions and can be denoted as $\vec{O}=[o^1, o^2, \dots, o^d] \in \R^d$. Consequently, the RUL prediction, $y$, at a given time $T$ is expressed as:
\begin{equation}
\label{eq:RUL_define}
y = f(\vec{X}_T;\vec{O}).
\end{equation}
Given a dataset, $\vec{D}=[(\vec{X}_T^1;\vec{O}^1;y^1), \dots, (\vec{X}_T^N;\vec{O}^N;y^N)]$, comprising $N$ sequences, our goal is to learn a mapping model to predict the RUL accurately.

\begin{figure}[htbp!]
\centering
  \begin{subfigure}{0.48\textwidth}
    \includegraphics[width=\textwidth]{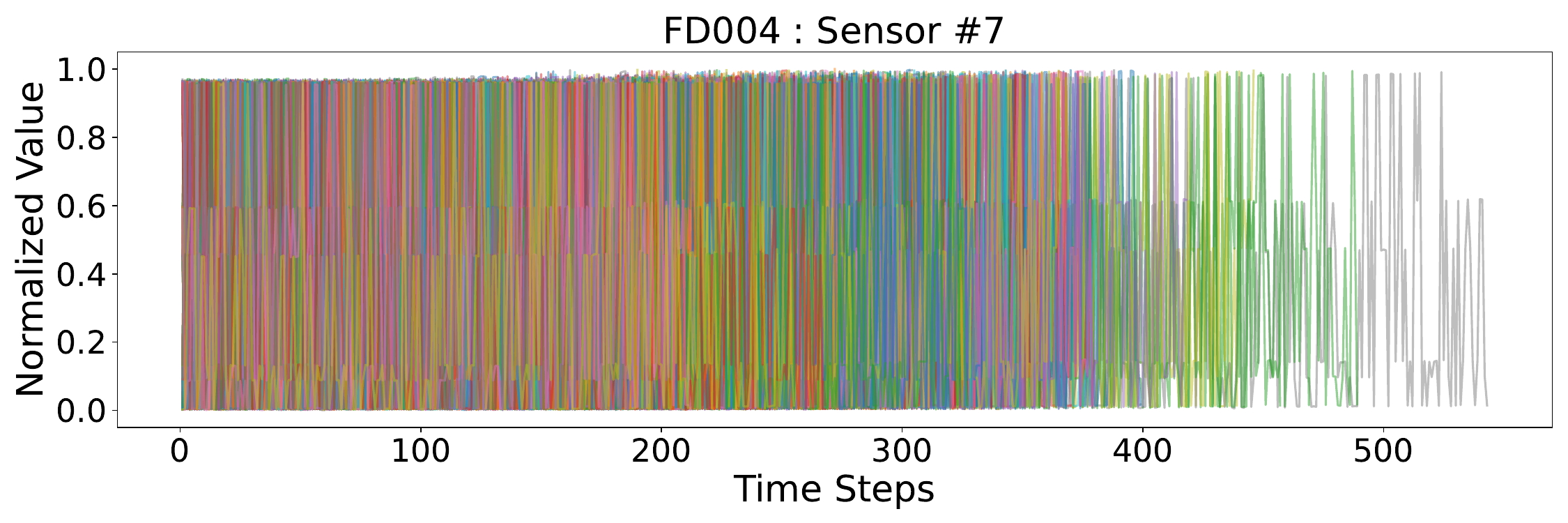}
      \caption{Unified Normalization}
      \label{subfig:Old_Normalization}
  \end{subfigure}
  \hfill
  \begin{subfigure}{0.48\textwidth}
    \includegraphics[width=\textwidth]{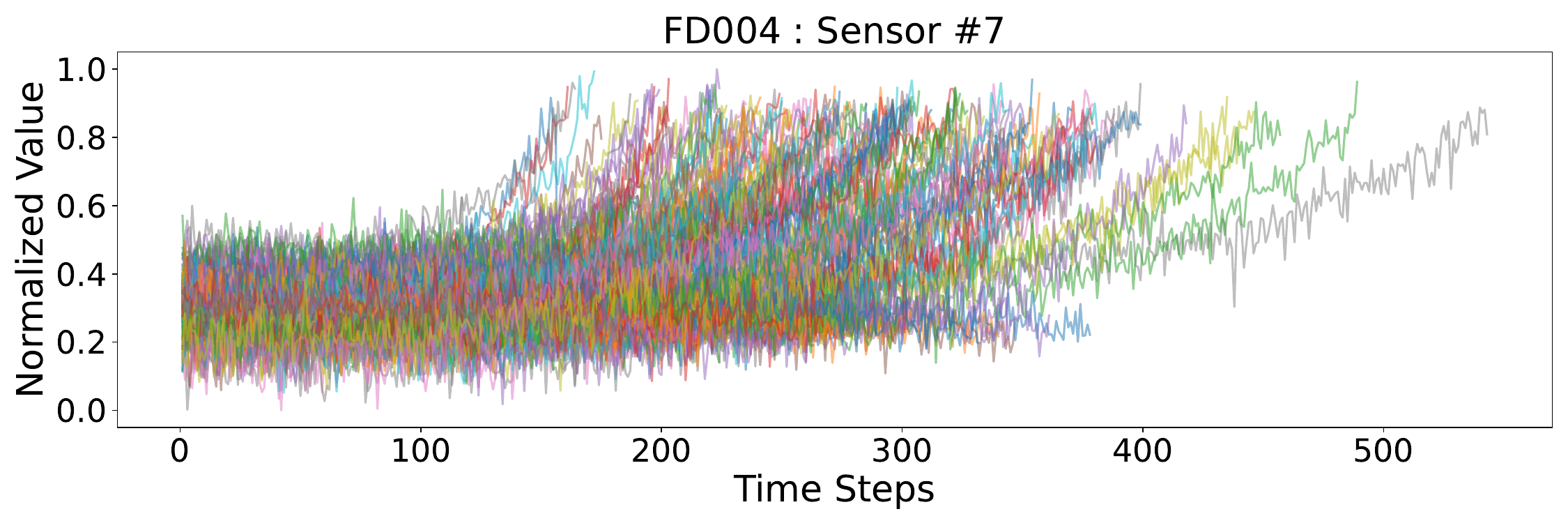}
      \caption{Clustering Normalization}
      \label{subfig:New_Normalization}
  \end{subfigure}
\caption{Normalization results based on two methods. Sequences with different colors represent the $N$ records of the 7th sensor.}
\label{fig:OP_Clustering_Normalization}%
\end{figure}

\subsection{Clustering Normalization based on Operating Conditions}
\label{subsec:clustering_normalization}



The disparity in the scale of data acquired by various sensors is often significant. Direct utilization of this data may skew the model's attention towards features of larger scales, neglecting those of smaller scales. Such an oversight can pose challenges during training and compromise prediction accuracy. Consequently, to counterbalance the influence of disparate feature scales, most RUL-related research undertakes the scaling of sensor data.
Let's denote $\vec{D}^i_{\vec{X}}$ as the all data from the $i$-th sensor of $\vec{X}_T$ within $\vec{D}$. The common method is unified normalization, i.e., the max-min normalization applied individually to each $\vec{D}^i_{\vec{X}} \in \R$, which scales the data within the interval $[0, 1]$. However, a single dataset often contains multiple operating conditions; that is, $\vec{O}^i \neq \vec{O}^j, \exists{i,j}$. As an illustrative example, the FD002 and FD004 datasets from C-MAPSS contain time series data spanning six operating conditions.

However, many existing works overlooked the impact of operating conditions during the preprocessing phase. Consider Fig~\ref{subfig:Old_Normalization}, where preprocessing of FD002 is executed using a unified normalization. We can observe that the time series, even under identical sensors, appears quite noisy, exhibiting pronounced fluctuations. Such inconsistencies can potentially impair the predictive performance of the RUL model.

A promising strategy involves clustering based on operating conditions, followed by individual normalization for each cluster. Let's represent the clustering model as $\mathbb{M}$ and designate $D_{\vec{O}}$ as $[\vec{O}^1,\dots,\vec{O}^N] \in \R^{d \times N}$, indicating the unique operating conditions associated with each time series. Upon clustering $\vec{D}_{\vec{O}}$, the output from $\mathbb{M}$ yields a label set $\{c_k^1,c_k^2,\dots,c_k^N\}$, where $c_k$ signifies the $k$-th cluster. A cluster, encompassing sequences sharing an identical cluster label, can be defined as:
\begin{equation}
\label{eq:cluster_data}
\vec{D}_{c_k}=[(X^l_T,O^l,y^l),...] \subseteq \vec{D}, \forall{l}~\mathbb{M}(O^l)= c_k.
\end{equation}
For each cluster, we perform max-min normalization to scale the data.
\begin{equation}
\label{eq:cluster_data}
X_{T_{norm}}^i = \frac{X_T^i - \min{D_{c_k}^i}}{\max{D_{c_k}^i} - \min{D_{c_k}^i}},
\end{equation}
where $i$ denote the $i$-th sensor, $X_T^i$ and $X_{T_{norm}}^i$ stand for raw and normalized time sequences of sensor $i$.
For the $i$-th sensor within cluster $c_k$, the maximum and minimum values are respectively denoted as $\max{D_{c_k}^i}$ and $\min{D_{c_k}^i}$.

The clustering normalized time series is depicted in Fig~\ref{subfig:New_Normalization}. When comparing with Fig~\ref{subfig:Old_Normalization}, it becomes evident that the time series, across varying operating conditions, manifest more consistent scales. We hypothesize that employing the clustering normalization approach can increase the RUL prediction accuracy of the model. Sec~\ref{subsec:clusering_effect} delves into the effects of cluster normalization on prediction, verified by comparative experimental results.

\subsection{Spatio-Temporal Attention Graph Neural Network}



The proposed STAGNN is primarily designed to extract the spatio-temporal relationships inherent in sensor sequences for RUL prediction. As illustrated in Fig~\ref{fig:STAGNN}, the model comprises three distinct modules: spatial feature extraction module, temporal feature extraction module, and prediction module.

\textbf{Spatial Feature Extraction Module}: This module employs GCN layers combined with a spatial attention mechanism. The graph is characterized by the equation $G = G(\vec{X}_T, \vec{A}, \vec{E})$. In line with Sec.~\ref{subsec:problem_statement}, $\vec{X}_T \in \R^{T \times S}$ signifies a multivariate sensor signal series where $T$ represents the time series length, and $S$ accounts for the number of sensors, considered as nodes within this graph representation. The set of edges in the graph is represented by $E$. The adjacency matrix, $\vec{A} \in R^{S \times S}$, reflects connections between sensors, with $\vec{A}{ij}=(v_i, v_j) \in E$ denoting a link between sensors $v_i$ and $v_j$. Given the absence of prior topological knowledge in data-driven RUL prediction methodologies, sensor relationships cannot be provided explicitly. Therefore, the adjacency matrix is formulated by assessing the covariance between sensors, as described by
\begin{equation}
\label{eq:adjacency_matrix}
\vec{A}_{ij} = \left\{\begin{matrix} 1, \text{if~} |cov(\vec{x}_T^i,\vec{x}_T^j)| > \lambda\\ 0, \text{if~} |cov(\vec{x}_T^i, \vec{x}_T^j)| < \lambda \end{matrix}\right . 
\end{equation}
where $\lambda$ is the threshold ranging from $0$ to $1$, a higher threshold implies that connections in the graph are reserved for sensors with closer relationships. 
We add a multi-head spatial attention layer after each GCN. This attention mechanism was first proposed by~\cite{velickovic2017graph}.

The output from a GCN layer, symbolized as  $\vec{H}_g \in \R^{n \times S}$, and $\vec{H}_g^i \in \R^n$ represents the sequence for $i$-th sensor. $\alpha_{ij}$ is the attention coefficient on sensors $v_i$ and $v_j$, is derived as:
\begin{equation}
\label{eq:attention}
\alpha_{ij} = \frac{exp\left(\text{LeakReLU}\left(\vec{W}_g^T\left[\vec{H}_g^i \parallel \vec{H}_g^j\right] \right)\right)}{\sum_{u \in \mathbb{N}_i}exp\left(\text{LeakReLU}\left(\vec{W}_g^T\left[\vec{H}_g^i \parallel \vec{H}_g^u\right]\right)\right)},
\end{equation}
where $\vec{W}_g  \in \R^{2n}$ is the weight matrix affiliated with the spatial attention layer,
$u \in \mathbb{N}_i$ means the neighbors of node $i$,
$\parallel$ stands for the concatenation operation and $\text{LeakReLU}$ is a nonlinear activation function.
Based on the calculated spatial attention coefficient, the $i$-th node's output of the multi-head spatial attention layer $\vec{H}_g^{i\text{'}}$ is defined as: 
\begin{equation}
\label{eq:attention}
\vec{H}_g^{i\text{'}} = \frac{1}{M}\sum_{m=1}^M\sum_{j \in \mathbb{N}_i} \alpha_{ij}^m\vec{H}_g^j 
\end{equation}
where $m$ enumerates the number of multi-head attention, and $\alpha_{ij}^m$ corresponds to the attention coefficient derived by the $m$-th attention head.

\textbf{Temporal Feature Extraction Module}: We incorporate TCN layers, which offer superior training efficiency compared to LSTM~\cite{bai2018empirical}. For temporal feature extraction, an effective strategy is to emphasize distinct regions of interest by allocating varied weights to time steps. Therefore, a multi-head temporal attention mechanism is appended to each TCN layer. This facilitates the discernment of the significance at different time steps.

Assume that the features procured from the TCN network for a given sample are represented as \( \vec{H}_t \in \mathbb{R}^{n \times S} \).
The importance vector, \( \vec{\beta} \in \mathbb{R}^{n} \), denotes the significance of one sensor across $n$ time steps and is expressed as:
\begin{equation}
\label{eq:temporal_attention}
\vec{\beta} = \frac{\exp(\text{Sigmod}(\vec{H}_t\vec{W}_t+\vec{b}))}{\sum \exp(\text{Sigmod}(\vec{H}_t\vec{W}_t+\vec{b}))},
\end{equation}
where \( \vec{W}_t \in \mathbb{R}^{S} \) and \( b \) are respectively the weight and bias parameters of the temporal attention layer. The resultant output from the multi-head temporal attention layer can be delineated as:
\begin{equation}
\label{eq:multi_head_output}
\vec{H}_t^{\text{'}} = \frac{1}{M}\sum_{m=1}^M \vec{H}_t \otimes \vec{\beta^m}.
\end{equation}
Here, \( M \) stands for the count of multi-head attentions, and \( \vec{\beta}^m \) embodies the attention coefficients for the \( m \)-th head, illustrated as \( \vec{\beta}^m = [\vec{\beta}_{1}^m, \vec{\beta}_{2}^m, \dots,\vec{\beta}_{S}^m] \). The operation \( \otimes \) is defined as an element-wise multiplication.

\textbf{Prediction Module}: This module fuses the extracted features. It mainly contains fully connected layers to get the final RUL prediction.
\section{Experiment}
\label{sec:experiment}
\subsection{Dataset Description}
\label{subsec:data_description}
The Commercial Modular Aerospace Propulsion System Simulation (C-MAPSS) dataset serves as the benchmark for assessing the efficacy of the proposed methodology. 
This dataset originates from an aircraft engine degradation simulation tool designed to emulate the engine degradation trajectory across various operational conditions. 
As delineated in Tab.~\ref{tab:dataset}, the dataset contains four subdatasets covering different operating conditions and is divided into training and testing subsets. Each record includes 26 features involving engine ID, operating time in cycles, 3 operating condition settings~($d$=3 in $\vec{O}$) and 21 sensors~($S$=21 in $\vec{X}_T$).

\begin{table}[htbp]
\caption{The details of C-MAPSS dataset}
\begin{center}
\begin{tabular}{ccccc}
\hline
Subsets & FD001 & FD002 & FD003 & FD004 \\ 
\hline
Number of Training Engines & 100 & 260 & 100 & 249 \\ 
Number of Test Engines & 100 & 259 & 100 & 248 \\ 
Number of Operation Conditions & 1 & 6 & 1 & 6 \\ 
\hline
\end{tabular}
\label{tab:dataset}
\end{center}
\end{table}

\subsection{Data Preprocess}
\label{subsec:data_preprocess}

Data preprocessing in the study is split into two principal stages: data normalization and data segmentation via sliding windows. The normalization procedure is described comprehensively in Sec.~\ref{subsec:clustering_normalization}. As shown in Fig.~\ref{fig:OP_Clustering_Normalization}, unified normalization might influence the prediction accuracy for RUL datasets across varying operational conditions. To discern the difference between the unified and clustering normalization, both techniques were employed to normalize the data. Subsequently, two types of normalized datasets were respectively trained on our proposed model with identical architecture. A detailed comparative analysis and results are described in Sec.~\ref{subsec:clusering_effect}.

In alignment with existing RUL prediction research, sliding windows are utilized to segment the primary data~\cite{heimes2008recurrent,zhang2023lstm}. Given a multivariate time series $ \vec{X}_T $ as detailed in Sec.~\ref{subsec:data_description}, where the sequence length is denoted by $ T $, we designate the window's length as $ w $ and the stride of the window as $ k $. This segmentation strategy splits $ \vec{X}_T $ into $ m $ subsequences, represented as $ \{\vec{X}_m\}_{m=1}^{\lfloor \frac{T-w}{k} \rfloor+1} $. Each subsequence's RUL, $y_m $, is defined as $ y_m = T-w-(m-1) \times k $. A piece-wise linear degradation model was implemented to cap the RUL's maximal value, with a threshold set at 125, as used by~\cite{heimes2008recurrent,zhang2023lstm}. The operating conditions and sensor measurements constituted the final input vectors for the prediction models.

\subsection{Evaluation Metrics}
\label{subsec:evaluation_metrics}
RMSE is widely used for RUL prediction as it is a common evaluation metric in regression tasks. RMSE is calculated as:
\begin{equation}
\label{eq:rmse}
\text{RMSE} = \sqrt{\frac{\sum_{i=0}^{N - 1} (y_i - \hat{y}_i)^2}{N}},
\end{equation}
where $N$ is the number of predicted samples, $y_i$ and $\hat{y}_i$ represent the actual RUL value and the predicted RUL value for the predicted sample respectively. In addition, Sateesh et. al~\cite{sateesh2016deep} proposed a scoring function and is widely used in RUL prediction research, which is defined as
\begin{equation}
\label{eq:score_function}
\text{Score} = \left\{\begin{matrix}\sum_{i=1}^{N}(e^{-\frac{\hat{y}_i-y_i}{13}}-1), \text{if~} \hat{y}_i < y_i\\\sum_{i=1}^{N}(e^{\frac{\hat{y}_i-y_i}{10}}-1), \text{if~} \hat{y}_i \geqslant y_i \end{matrix}\right .
\end{equation}
Unlike RMSE which can not distinguish between early and delayed predictions, the scoring function will bring more penalties to delayed RUL predictions, which is more in line with actual application scenarios. In this article, we used these two criteria to evaluate our proposed model's performance.

\subsection{Experiment Results}
\label{subsec:experiment_results}
\begin{figure}[htbp!]
\centering
  \begin{subfigure}{0.5\textwidth}
    \includegraphics[width=\textwidth]{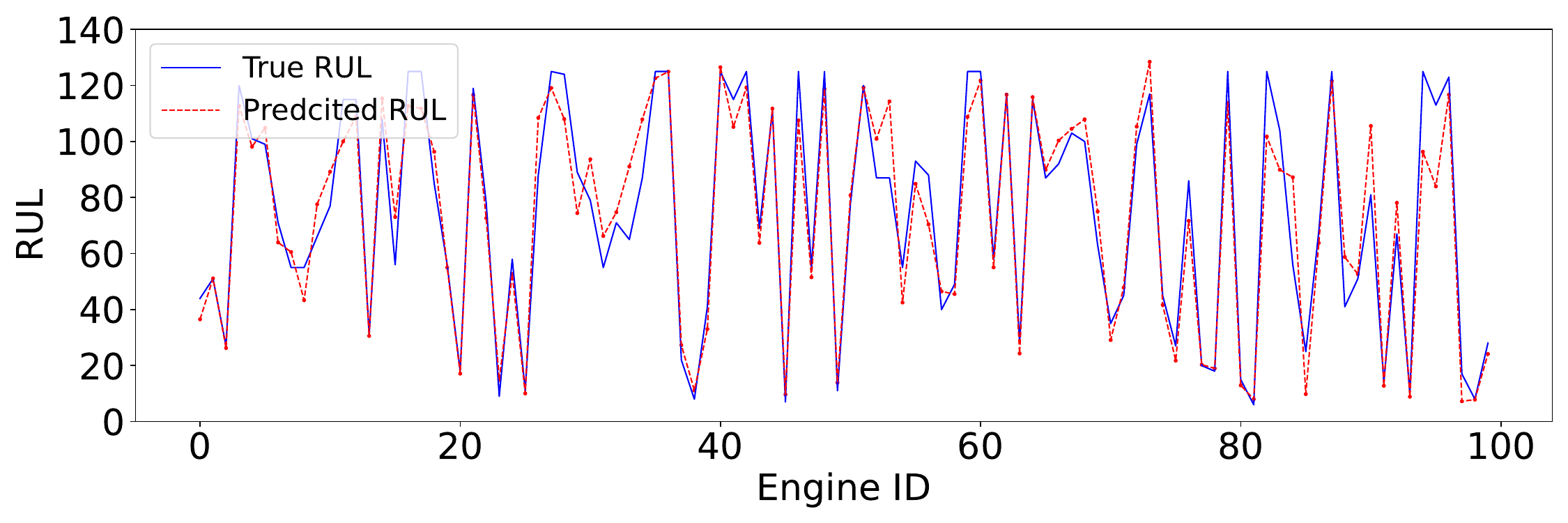}
      \caption{Prediction Result of STAGNN* on Test Dataset of FD003 }
      \label{subfig:FD003}
  \end{subfigure}
  \hfill
  \begin{subfigure}{0.5\textwidth}
    \includegraphics[width=\textwidth]{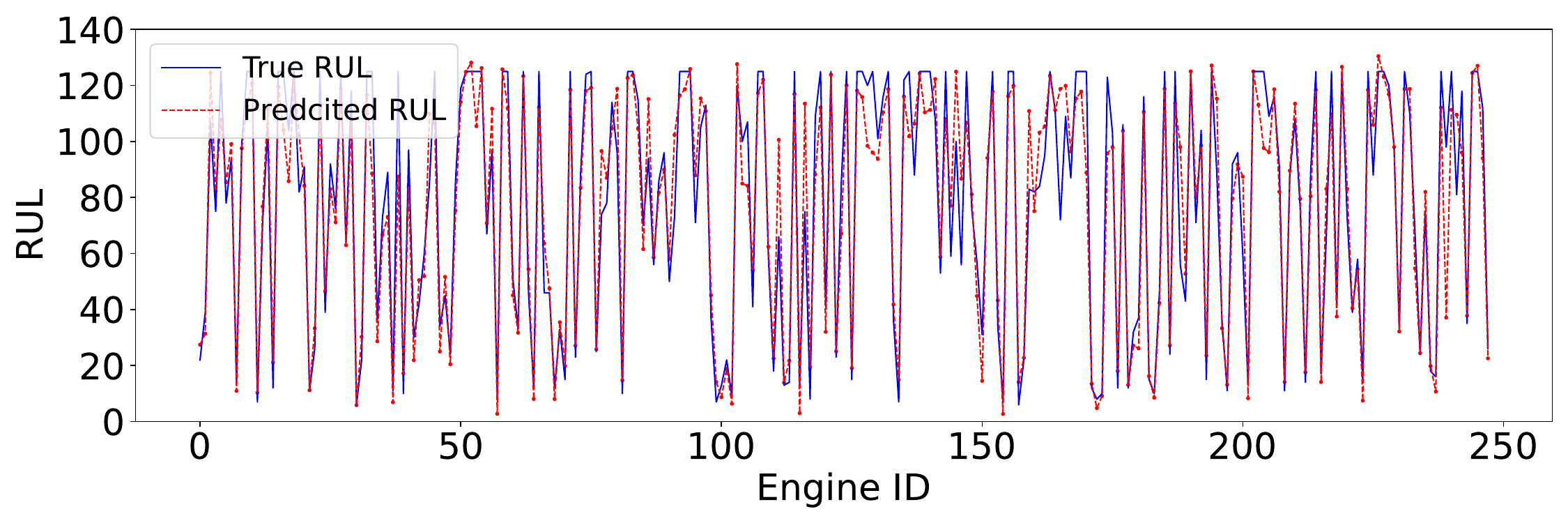}
      \caption{Prediction Result of STAGNN* on Test Dataset of FD004 }
      \label{subfig:FD004}
  \end{subfigure}
\caption{True RUL and the predicted RUL by the proposed approach on the two test datasets. }
\label{fig:Prediction_Visualization}%
\end{figure}

As illustrated in Fig.~\ref{fig:STAGNN}, our model incorporates a two-layer Attention Graph Neural Network (AGNN) with a hidden feature dimension of 64. The AGNN, which is responsible for spatial feature extraction, feeds its output into an attention-based TCN block (ATCN) designed for temporal feature extraction. This ATCN comprises two layers of TCN-Attention structures with hidden dimensions of 64 and 10, respectively. The ATCN has a kernel size of two and employs a dropout rate of 0.5. Its output serves as the feature representation derived from the hidden layers. To delve into the model's explainability, we visualize this hidden feature representation in Sec.~\ref{subsec:explainability_exploration}. Subsequently, this representation is channeled through a fully connected layer to get RUL predictions.

In line with the methodology presented in~\cite{liu2020remaining}, we employ a sliding window of length $w=50$ for data segmentation, with a step size $k=1$. The Adam optimizer is utilized with a learning rate set to $0.001$. Our model is trained using batches of size 100 over 100 epochs. For each dataset, we conducted ten trials and subsequently computed the mean values for both RMSE and Score.

\begin{table}[htbp]
\caption{Performance Comparison on C-MAPSS dataset}
\setlength{\tabcolsep}{0.4mm}{
\begin{center}
\begin{tabular}{ccccccccc}
\hline
 & \multicolumn{2}{c}{FD001} & \multicolumn{2}{c}{FD002} & \multicolumn{2}{c}{FD003} & \multicolumn{2}{c}{FD004} \\ 
Approaches&RMSE&Score&RMSE&Score&RMSE&Score&RMSE&Score\\
\hline
DCNN~\cite{li2018remaining} & 12.61 & 273.7 & 22.36 & 10412 & 12.64 & 284.1 & 23.31 & 12466 \\
HALSTM~\cite{chen2020machine} & 14.53 & 322.4 & N/A& N/A &N/A& N/A &27.08& 5649.1\\
RNN~\cite{yu2020improved} & 13.58 & 228& 19.59 &2650 &19.16 &1727 &22.15 &2901 \\
AGCNN~\cite{liu2020remaining} & 12.42 & 225.5 & 19.43 & 1492 & 13.39 & 227 & 21.50 & 3392 \\
Transformer~\cite{mo2021remaining} & \textbf{11.27} & N/A & 22.81 & N/A & 11.42 & N/A & 24.86 & N/A \\
GCN~\cite{wang2021spatio} & 12.76 & 266 & N/A & N/A & 12.07 & 278 & N/A & N/A \\
LSTMBS~\cite{liao2018uncertainty} &  14.89 &  481.1 &  26.86 &  7982 &  15.11 & 493.4 &  27.11 & 5200 \\
\textbf{STAGNN} & 11.50 & \textbf{187.2} & \textbf{16.92} & \textbf{1132.2} & \textbf{11.05} & \textbf{196.0} & \textbf{19.33} & \textbf{1443.4} \\
\hline
DLSTM*~\cite{miao2019joint} & 12.29 & N/A & 17.87 & N/A & 14.34 & N/A & 21.81 & N/A \\
HAGCN*~\cite{li2021hierarchical} & 11.93 & 222.3 & 15.05 & 1144.1 & 11.53 & 240.3 & 15.74 & 1218.6 \\
BDL*~\cite{lin2022bayesian} & 18.6 & 2774 & 22.9 & 7734 & 27.9 & 19990 & 28.1& 53295 \\
GAT*~\cite{li2022emerging} & 13.21 & 303.1 & 17.25 & 5338.8 & 15.36 & 507.5 & 21.44 & 2971.9 \\
RGCNU*~\cite{zhu2023rgcnu} & \textbf{11.18} & \textbf{173.5} & 16.22 & 1148.16 & 11.52 & 225.0 & 19.11 & 2215.9 \\
\textbf{STAGNN*} & 11.50 & 187.2 & \textbf{13.81} & \textbf{826.3} & \textbf{11.05} & \textbf{196.0} & \textbf{14.30} & \textbf{1038.5} \\
\hline
\end{tabular}
\label{tab:result}
\end{center}
}
\end{table}

Tab.~\ref{tab:result} presents the performance outcomes of recent RUL-related studies. As highlighted in Sec~\ref{sec:intro}, previous research has employed various data preprocessing techniques. To ensure a fair comparison, we have categorized the results into two primary groups: unified normalization and cluster normalization. The asterisk (*) denotes methods utilizing cluster normalization. While RGCNU~\cite{zhu2023rgcnu} does not specify its preprocessing approach, based on its referenced literature, we placed it under the cluster normalization category. STAGNN and STAGNN* represent the performances of our proposed method within the respective groups. Notably, STAGNN demonstrates SOTA performance even in the absence of cluster normalization. Our method's RMSE for FD001 is slightly lower than that of Transformer~\cite{mo2021remaining} and RGCNU*~\cite{zhu2023rgcnu}, but surpasses these two methods in more complex datasets such as FD002 and FD004 by up to 25.8\%. This underscores our method's robustness in predicting RUL under complex operating conditions. These comparative insights address \textbf{question~1} from Sec.~\ref{sec:intro}, confirming that a model incorporating a cascaded spatio-temporal feature extractor can also achieve SOTA results with a more streamlined structure than another hybrid model by~\cite{wang2021spatio}.

\subsection{Analysis the Effectiveness of Clustering Normalization}
\label{subsec:clusering_effect}

Utilizing cluster normalization in the data preprocessing phase significantly enhances the RUL prediction performance of STAGNN* on the FD002 and FD004 datasets, which contain multiple operating conditions. As shown in Tab.~\ref{tab:result}, RMSE and Score achieved improvements of up to 26\% and 27\%, respectively. These findings affirm \textbf{question~3} from Sec.~\ref{sec:intro}: cluster normalization effectively mitigates the influence of varied operating conditions on model predictions. Although numerous studies, as indicated in Tab.~\ref{tab:result}, focus on end-to-end models aiming for performance boost through model refinements, the pivotal role of feature engineering is often overlooked.

\subsection{Model Ablation Study}
\label{subsec:model_ablation_study}
\begin{table}[htbp]
\caption{Performance Comparison for Model Ablation}
\setlength{\tabcolsep}{0.6mm}{
\begin{center}
\begin{tabular}{ccccccccc}
\hline
 & \multicolumn{2}{c}{FD001} & \multicolumn{2}{c}{FD002} & \multicolumn{2}{c}{FD003} & \multicolumn{2}{c}{FD004} \\ 
Approaches&RMSE&Score&RMSE&Score&RMSE&Score&RMSE&Score\\
\hline
GNN & 13.07 & 276.6 & 17.95 & 1545.8 & 11.55 & 207.8 & 20.66 & 2265.4 \\
AGNN & 12.61 & 270.1 & 17.91& 1244.1 &11.31& 204.2 &20.08& 2069.1\\
TCN & 11.63 & 199.8& 22.09 &2968.8 &11.18 &208.7 &23.42 &3020.4 \\
ATCN & 11.75 & 218.4 & 21.88 & 2193.9 & 13.34 & 249.7 & 22.18 & 2436.4 \\
STGNN & 11.55 & 192.6 & 17.63 & 1314.3 & 11.18 & 207.1 & 20.05 & 1771.1 \\
STAGNN & \textbf{11.50} & \textbf{187.2} & \textbf{16.92} & \textbf{1132.2} & \textbf{11.05} & \textbf{196.0} & \textbf{19.33} & \textbf{1443.4} \\

\hline
\end{tabular}
\label{tab:model_ablation}
\end{center}
}
\end{table}

Ablation analysis of the model aids in evaluating the contribution of each component to the RUL prediction. We decomposed the proposed STAGNN model into five distinct sub-models for comparative assessment. These are: a two-layer GNN dedicated solely to spatial feature extraction; an AGNN, which is a GNN enhanced with spatial attention; a two-layer TCN focusing exclusively on temporal feature extraction; an ATCN, which is a TCN augmented with temporal attention; and a cascaded GNN-TCN model devoid of any attention mechanism, termed STGNN. Unified normalization was applied during data preprocessing to assess the model's adaptability to complex datasets. 

We discern three key observations based on Table~\ref{tab:model_ablation}. Firstly, while the GNN surpasses the TCN by a margin of up to 25\% in terms of Score on intricate datasets FD002 and FD004, the outcomes are inversed for FD001 and FD003. This suggests that spatial features become more influential for RUL predictions on complex datasets. Secondly, the cascaded STGNN consistently outperforms standalone GNN and TCN models, particularly in terms of Score. This underscores the significance of both spatial and temporal features in RUL prediction. Lastly, comparing models with and without attention mechanisms reveals that incorporating attention increases RUL prediction accuracy on complex datasets, addressing \textbf{question~2} from Sec.~\ref{sec:intro}. The attention mechanism also potentially enhances model explainability, briefly analyzed in Sec.~\ref{subsec:explainability_exploration}.


\subsection{Explainability Exploration}
\label{subsec:explainability_exploration}
Deep neural networks are powerful but often operate as black boxes, making their decision-making processes opaque to humans. Consequently, model explainability has gained prominence. Visualizing both the hidden feature representation and the attention matrix might offer insights into dissecting model decisions. 
\begin{figure}[htbp!]
\centering
  \begin{subfigure}{0.24\textwidth}
    \includegraphics[width=\textwidth]{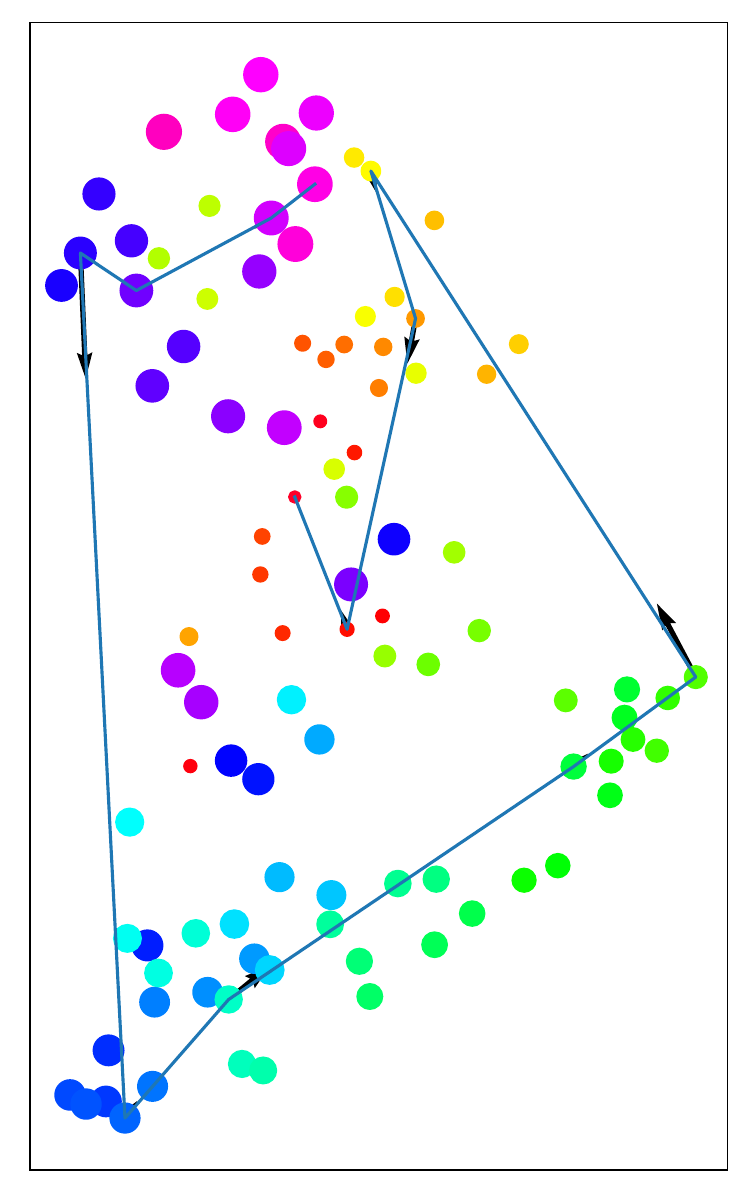}
      \caption{STAGNN}
      \label{subfig:STAGNN}
  \end{subfigure}
  \hfill
  \begin{subfigure}{0.24\textwidth}
    \includegraphics[width=\textwidth]{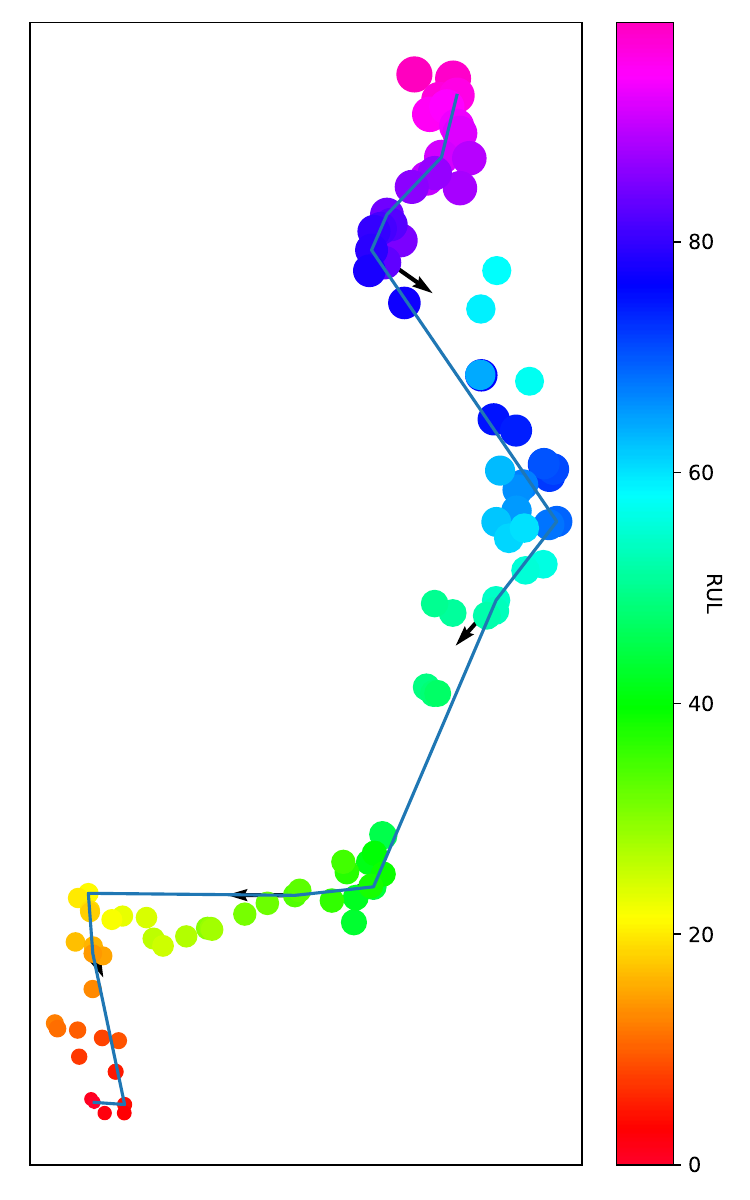}
      \caption{STAGNN*}
      \label{subfig:STAGNN*}
  \end{subfigure}
\caption{The t-SNE-based visualization of the feature representations for STAGNN and STAGNN*. The color and size represent the degradation of RUL}
\label{fig:t-SNE_Visualization}%
\end{figure}

Fig.~\ref{subfig:STAGNN} and Fig.~\ref{subfig:STAGNN*} present a t-SNE-based 2D visualization of the feature representations for STAGNN and STAGNN*, respectively, under the FD002 dataset. Each plotted point corresponds to the model's feature representation at a specific RUL, with the point's color and size representing the degradation of RUL.
In Fig.~\ref{subfig:STAGNN}, the feature representation for STAGNN follows a spiral trajectory as the RUL decreases. Notably, several points deviate from this trajectory. While STAGNN can adapt to noisy data, it is not entirely immune to noise-induced prediction. In contrast, the trajectory in Fig.~\ref{subfig:STAGNN*}, representing STAGNN*, is more distinct. Observing Fig.~\ref{subfig:STAGNN*} closely, we note significant jumps when the RUL is approximately 70 and 35. This suggests potential abrupt changes in a component state around 70 and 35, which could lead to rapid performance degradation of the entire system. 
\begin{figure}[htbp!]
\centering
  \begin{subfigure}{0.24\textwidth}
    \includegraphics[width=\textwidth]{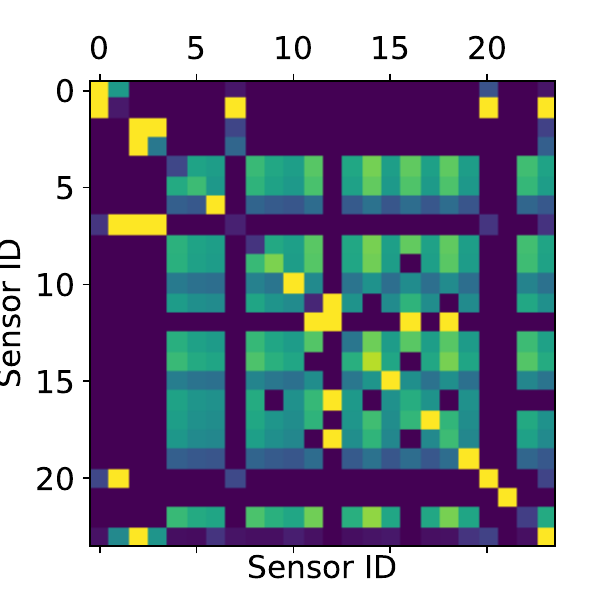}
      \caption{Spatial Attention}
      \label{subfig:spatial_att}
  \end{subfigure}
  \hfill
  \begin{subfigure}{0.24\textwidth}
    \includegraphics[width=\textwidth]{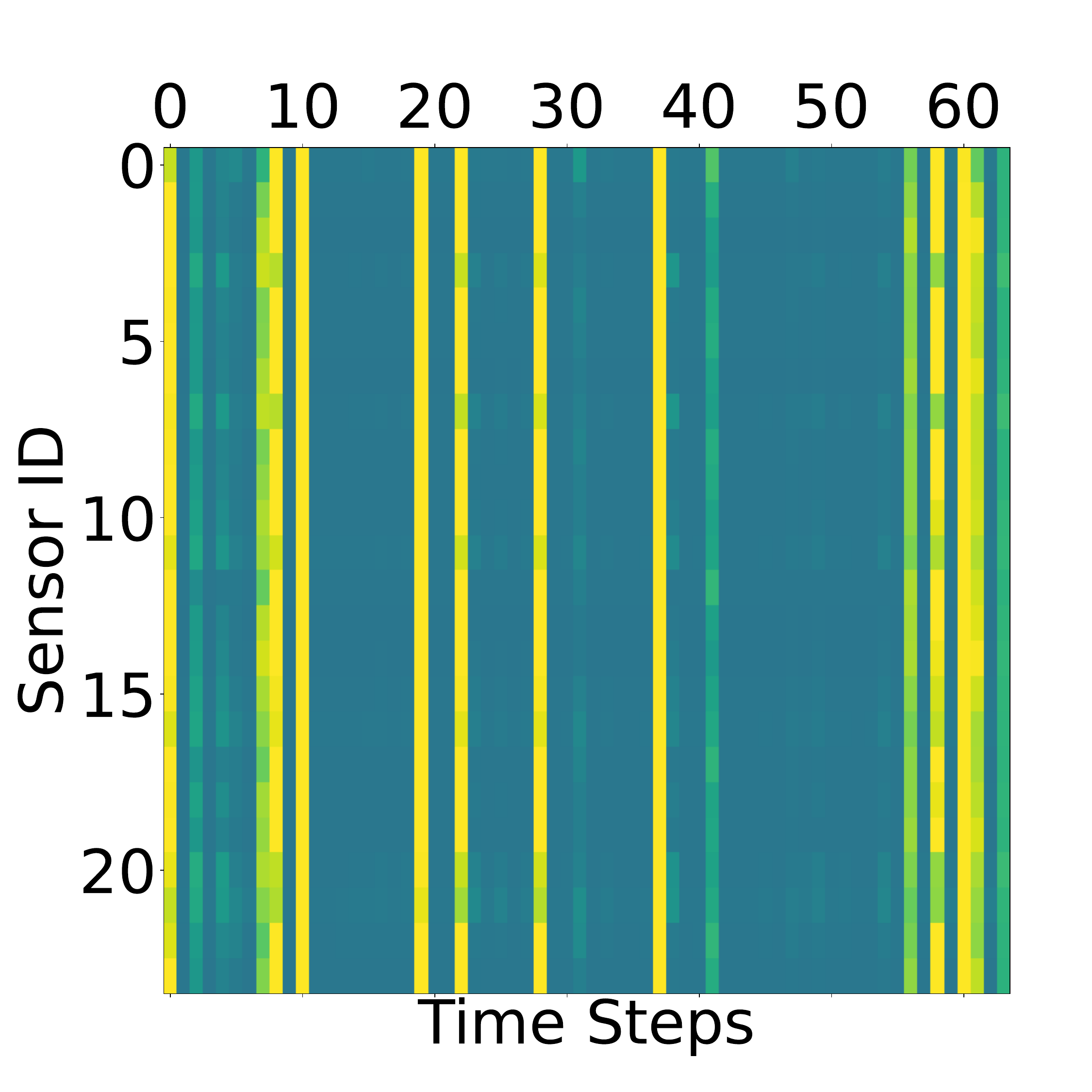}
      \caption{Temporal Attention}
      \label{subfig:temporal_att}
  \end{subfigure}
\caption{Samples of spatial and temporal attention matrix. }
\label{fig:attention_Visualization}%
\end{figure}

Fig.~\ref{subfig:spatial_att} and Fig.~\ref{subfig:temporal_att} present visualizations of the spatial and temporal attention matrix, respectively. The spatial attention matrix illustrates the attention between sensors. A large weight indicates a stronger connection between the components denoted by the two sensors, suggesting a more pronounced influence on the RUL. In addition, the temporal attention weight matrix demonstrates the impact of a sensor on the RUL at specific time steps in the series. Such insights could guide human experts in identifying root causes and directing maintenance efforts toward relevant system components.

\section{Conclusion}
\label{sec:conclusion}

In this study, we address three questions related to RUL prediction: 1) the efficacy of a cascading spatio-temporal feature extractor structure; 2) the influence of the attention mechanism on prediction accuracy; and 3) the effect of operating conditions on RUL prediction.

We introduced STAGNN, a cascading spatio-temporal feature extractor that integrates GNN and TCN. This model also incorporates tailored attention mechanisms corresponding to different feature extractors. 
Our results demonstrated SOTA performance in comparison to existing works. Furthermore, by employing clustering normalizing based on operating conditions, we significantly enhanced the RUL prediction accuracy in datasets containing diverse operating conditions. We visualized the hidden layer feature representation and the attention matrix to explore the model transparency.

Our future work will encompass evaluating prediction uncertainties using Bayesian neural networks and delving into techniques for model explainability, aiming to quantify the explanation result.

\bibliographystyle{IEEEtran}
\bibliography{conference_101719}

\end{document}